\documentclass[final]{cvpr}

\usepackage{times}
\usepackage{epsfig}
\usepackage{graphicx}
\usepackage{amsmath}
\usepackage{amssymb}
\usepackage{microtype}      
\usepackage{mathrsfs}  
\usepackage{bm}
\usepackage{amssymb,amsmath}
\usepackage{multicol,multicap,graphicx}
\usepackage{graphicx}
\usepackage{subcaption}
\usepackage{wrapfig}
\usepackage{picinpar}
\usepackage{cutwin}
\usepackage{stmaryrd}
\usepackage{graphicx,multicol}
\usepackage{algorithm}  
\usepackage{algorithmic}
\usepackage{amsfonts}
\usepackage{amsfonts,amssymb} 
\usepackage{verbatim}
\usepackage{url}
\usepackage{enumitem}

\DeclareMathOperator*{\argmin}{arg\,min}

\usepackage{booktabs}
\usepackage{multirow}
\DeclareMathOperator{\Exp}{Exp}
\DeclareMathOperator{\Log}{Log}

\usepackage[pagebackref=true,breaklinks=true,colorlinks,bookmarks=false]{hyperref}

\def\cvprPaperID{8} 
\def\confYear{DiffCVML 2021}

\begin{document}

\title{Deep Spherical Manifold Gaussian Kernel for Unsupervised Domain Adaptation}

\author{Youshan Zhang \ \ \ \ \ \ \ \  Brian D.\ Davison \\
Computer Science and Engineering, Lehigh University, Bethlehem, PA, USA\\
{ \{yoz217, bdd3\}@lehigh.edu}
}

\maketitle
\thispagestyle{empty}
\pagestyle{empty}

\begin{abstract}
Unsupervised Domain adaptation is an effective method in addressing the domain shift issue when transferring knowledge from an existing richly labeled domain to a new domain. Existing manifold-based methods either are based on traditional models or largely rely on Grassmannian manifold via minimizing differences of single covariance matrices of two domains. In addition, existing pseudo-labeling algorithms inadequately consider the quality of pseudo labels in aligning the conditional distribution between two domains. In this work, a deep spherical manifold Gaussian kernel (DSGK) framework is proposed to map the source and target subspaces into a spherical manifold and reduce the discrepancy between them by embedding both extracted features and a Gaussian kernel. To align the conditional distributions, we further develop an easy-to-hard pseudo label refinement process to improve the quality of the pseudo labels and then reduce categorical spherical manifold  Gaussian kernel geodesic loss.  Extensive experimental results show that DSGK outperforms state-of-the-art methods, especially on challenging cross-domain learning tasks.
\end{abstract}

\section{Introduction}
Massive amounts of labeled data are a prerequisite of most existing machine learning methods. Unfortunately, such a requirement cannot be met in many real-world applications. In addition, collecting sufficient labeled data is a big investment of time and effort. Therefore, it is often necessary to transfer label knowledge from one labeled domain to an unlabeled domain. However, due to domain shift or dataset bias issue~\cite{pan2010survey}, it is difficult to improve performance for the unlabeled domain. 

Domain adaptation (DA) is proposed to circumvent the domain shift problem. By not requiring additional annotated labels on the new domain, unsupervised DA (UDA) is attractive, as it aims to transfer knowledge learned from a label-rich source domain to a fully unlabeled target domain~\cite{long2016unsupervised}. Before the popularity of deep features, approaches with hand-crafted features aim to map the two domains into a shared subspace and learn the invariant features~\cite{zhang2020domain}. Manifold learning is  commonly used to identify the shared space between the source and target domains. There have been efforts made in traditional methods, including sampling geodesic flow (SGF)~\cite{gopalan2011domain}, geodesic flow kernel (GFK)~\cite{gong2012geodesic}, and geodesic sampling on manifolds (GSM)~\cite{zhang2019transductive}. These methods focus on matching either marginal, conditional, or joint distributions between two domains to learn domain-invariant representations. However, these traditional methods cannot handle large-scale recognition tasks since they require large memory to compute singular value decomposition (SVD) on a Grassmannian manifold. Although discriminative manifold propagation (DMP)~\cite{luo2020unsupervised} proposed a Grassmannn distance to reduce the domain discrepancy,  it still largely relies on the differences of covariance matrices, and it cannot avoid complex the SVD process.
Further, its Log-Euclidean loss is not the closed-form solution to calculate the intrinsic distance between two domains. Recently, existing deep learning-based methods generate pseudo labels for the target domain to align the conditional distribution and learn the target discriminative representations~\cite{xie2018learning,zhang2018collaborative}. However, the credibility of these pseudo labels is unknown. Noisy labels can easily lead to poor alignment and discrimination. As a result, it is easy to cause negative transfer for the target domain.

To address the above challenges, our contributions are three-fold:
\begin{itemize}[noitemsep,topsep=0pt]
    \item To explicitly measure the intrinsic distance between two domains and reduce computation time, we propose a novel spherical manifold  Gaussian kernel geodesic loss, which considers both latent features and discrepancy between covariance matrices.
    
    \item We develop an easy-to-hard refinement process to remove the noise labels via $T$ times adjustment, and we then form a pseudo labeled target domain so as to jointly optimize the shared classifier with labeled examples from the source domain.
    
    \item We also enforce a categorical spherical manifold  Gaussian kernel geodesic loss to reduce conditional discrepancies. Then, our model can jointly align marginal and conditional distributions between two domains.
    
\end{itemize}
We conduct extensive experiments on three benchmark datasets (Office-31, Office-Home, and VisDA-2017), achieving higher accuracy than state-of-the-art methods.

\section{Related work}
Most existing manifold-based methods are only focused on the Grassmannian manifold. The SGF model~\cite{gopalan2011domain} generated multiple intermediate subspaces between the source and the target domain along with geodesic flow on a Grassmannian manifold. Then, GFK~\cite{gong2012geodesic} integrated all sampled points along the geodesic as calculated in the SGF model via constructing a kernel function. Manifold embedded distribution alignment (MEDA)~\cite{wang2018visual} further took the advantages of GFK to better represent source and target domain features and then dynamically aligned the marginal and conditional distributions between two domains. Later, geodesic sampling on manifolds (GSM)~\cite{zhang2019transductive} revealed that the SGF model cannot generate correct intermediate subspaces along the true geodesic, and provided a correct way to sample the intermediate features along the correct geodesic for the general manifold, which is further extended to the sphere, Kendall's shape, and Grassmannian manifold. Luo et al.~\cite{luo2020unsupervised} proposed a discriminative manifold propagation for UDA in a deep learning framework. They proposed Grassmann distance and the Log-Euclidean loss to minimize the difference between the two domains. They did not, however, explicitly measure the intrinsic distance between two domains. 

Other frequently used deep learning-based methods rely on minimizing the discrepancy between the source and target distributions by proposing different loss functions, such as Maximum Mean Discrepancy (MMD)~\cite{tzeng2014deep}, CORrelation ALignment~\cite{sun2016deep}, and Kullback-Leibler divergence~\cite{meng2018adversarial}.  Recently, 
Kang et al.~\cite{kang2019contrastive} extended MMD to the contrastive domain discrepancy loss. Li et al.~\cite{li2020enhanced} proposed an Enhanced Transport Distance (ETD) to measure domain discrepancy by establishing the transport distance of attention perception as the predictive feedback of iterative learning classifiers.
However, these distance-based metrics can also mix samples of different classes together. Inspired from GAN~\cite{goodfellow2014generative}, adversarial learning has shown its power in learning domain invariant representations. The domain discriminator aims to distinguish the source domain from the target domain, while the feature extractor aims to learn domain-invariant representations to fool the domain discriminator~\cite{ganin2016domain,tzeng2017adversarial,zhang2021adversarial1}. Pseudo-labeling is another technique to address UDA and also achieves substantial performance on multiple tasks. There are also many methods that utilized pseudo-labels to consider label information in the target domain and then minimize the conditional distribution discrepancy between two domains~\cite{saito2017asymmetric,zhang2018collaborative,kang2019contrastive,zhang2021adversarial2}. However, it is still difficult to remove  noisy pseudo labels for the target domain. Notably, we project data into a much faster spherical manifold and propose a useful easy-to-hard refinement process.

\section{Methodology}
\vspace{-0.1cm}
\subsection{Problem}
Here we discuss the unsupervised  domain adaptation (UDA) problem  and introduce some basic notation. Given a labeled source domain $\mathcal{D_S} = \{\mathcal{X}_\mathcal{S}^i, \mathcal{Y}^i_\mathcal{S} \}_{i=1}^\mathcal{N_S}$ with $\mathcal{N_S}$ samples in $C$ categories and an unlabeled target domain $\mathcal{D_T} = \{\mathcal{X}_\mathcal{T}^j\}_{j=1}^{\mathcal{N_T}}$ with $\mathcal{N_T}$ samples in the same $C$ categories  ($\mathcal{Y_T}$ for evaluation only), our challenge is how to get a well-trained classifier so that domain discrepancy is minimized and generalization error in the target domain is reduced.

In UDA, existing manifold-based methods are either based on traditional methods~\cite{gopalan2011domain,gong2012geodesic,zhang2019transductive} or highly rely on the Grassmannian manifold, which requests complex singular value decomposition (SVD) of the covariance matrices~\cite{luo2020unsupervised}. In addition, to align the conditional distributions of two domains, the reliability of generated pseudo labels is uncertain. These approaches face two critical limitations: (1) the SVD needs more computation time, and minimizing covariance matrices is not equivalent to reducing marginal distribution differences between two domains. It is hence necessary to develop a faster metric to align the marginal distribution of two domains. (2)  Noisy pseudo labels can deteriorate the shared classifier. The categorical condition distribution of two domains is difficult to minimize with the lower quality pseudo labels. 

To mitigate these shortcomings, we propose a deep spherical manifold Gaussian kernel (DSGK) model. To avoid the complex calculation of SVD, we focus on a spherical manifold. We propose a spherical manifold Gaussian kernel geodesic loss to minimize the marginal distribution, and design an easy-to-hard pseudo-label refinement process to improve the quality of the pseudo-labels in the target domain and then minimize the categorical spherical manifold Gaussian kernel geodesic loss. Therefore, our DSGK model can jointly align the marginal and conditional distributions of two domains.

\subsection{Geometry of spherical manifold}
\begin{figure}[t]
\centering
\includegraphics[width=0.8\columnwidth]{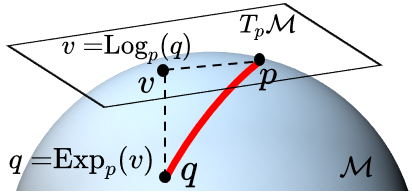}
\caption{Some basic concepts of geometry on manifold $\mathcal{M}$. $p$ and $q$ are two points on $\mathcal{M}$. $T_{p}\mathcal{M}$ is the tangent space at point $p$ and $p$ is called the pole of the tangent space. The red curve $\gamma$ is called the geodesic, which is shortest distance between $p$ and $q$ on $\mathcal{M}$. The Logarithmic map $\Log_p (\cdot)$ projects the point $p$ into the tangent space and the Exponential map $\Exp_p (\cdot)$ projects the element of tangent space $v$ back to the manifold, such that $v = \Log_p(q)$ and $\Exp_p(v) = \gamma(1) = q$.
}
\vspace{-0.3cm}
\label{fig:mani}
\end{figure}

Before, we discuss the contributions of this work, we first recap some important concepts on Riemannian manifold as shown in Fig.~\ref{fig:mani}. The $n$ dimensional unit sphere denoted as $S^n$ and can be defined as $S^n = \{ (x_1, x_2, \cdots, x_{n+1}) \in \mathbb{R} ^{n+1} | \sum_{i=1}^{n+1} x_i^2 =1 \}$.  Let $p$ and $q$ be two points on a sphere $S^n$ embedded in $\mathbb{R}^{n+1}$, and the tangent space of $S^n$ at point $p$ is denoted as $T_pS^n$.

The Logarithmic ($\Log$) map between $p$ and $q$ can be computed in Eq.~\ref{eq:Log_Sphere}.
\begin{equation}\label{eq:Log_Sphere}
\begin{aligned}
    & v =\text{Log}(p,q)=\frac{\theta \cdot L}{||L||}, \quad 
    \theta=\arccos ( \left< p,q \right>), \quad  \\ & L=(q-p\cdot \left< p, q \right>)
\end{aligned}
\end{equation}
where $p\cdot \left< p, q \right> $ denotes the projection of the vector $q$ onto $p$. The norm of $v$ is usually a constant, $\textit{i.e.}$ $||v|| = const.$, which is the distance between point $p$ and $q$. 

Given point $p$, and its tangent vector $v$ from Eq.~\ref{eq:Log_Sphere} and $t$, the  Exponential ($\Exp$) map is defined as:
\begin{equation}\label{eq:Exp_Sphere}
    \text{Exp}(p, vt)=\cos{\theta} \cdot p +\frac{\sin{\theta}}{\theta}\cdot vt,\  \theta= ||vt||.
\end{equation}
Additional details of $\Log$ map and $\Exp$ map on the spherical manifold can be found in~\cite{wilson2010spherical,zhang2019mixture,zhang2019k}. 

\begin{figure*}[h]
\centering
\includegraphics[width=2\columnwidth]{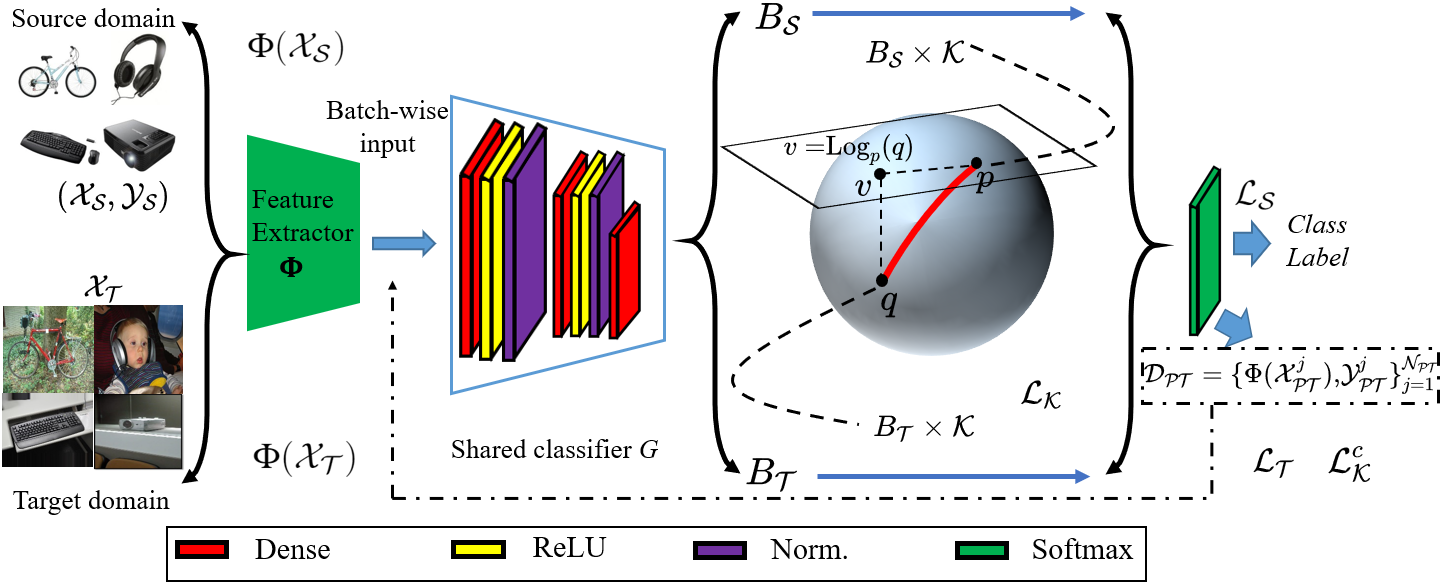}
\caption{Architecture of the DSGK model. We first extract features $\Phi(\mathcal{X_{Z}})$ for both source and target domains via $\Phi$ using a pre-trained model, and then train the shared classifier $G$. $\mathcal{L_S}$ is source classification loss.  Spherical manifold Gaussian kernel geodesic loss $\mathcal{L_{K}}$ minimize the marginal distribution difference of two domains. The dash-dot line is the generated pseudo labeled target domain using an easy-to-hard refinement process, which will optimize the shared classifier $G$ in $T$ times. $\mathcal{L_T}$ is the pseudo labeled target domain classification loss, and $\mathcal{L}_\mathcal{K}^c$ is categorical spherical manifold Gaussian kernel geodesic loss to minimize the conditional distribution. Norm.: BatchNormalization layer.  Best viewed in color.}  
\vspace{-0.3cm}
\label{fig:model}
\end{figure*}

\subsection{Initial source classifier}
Given feature activation function $\Phi$ from one backbone network, it maps data into a $d$ dimensional latent space, \textit{i.e.,} $\Phi(\mathcal{X_Z}) \in \mathbb{R}^{\mathcal{N_Z} \times d}$, where $\mathcal{Z}$ can be either source domain $\mathcal{S}$  or target domain $\mathcal{T}$.
The task in the source domain is trained using the typical cross-entropy loss as follows:
\begin{equation}\label{eq:lc}
    \mathcal{L_S} = - \frac{1}{\mathcal{N}_\mathcal{S}}\sum_{i=1}^{\mathcal{N}_\mathcal{S}} \sum_{c=1}^{C} \mathcal{Y}_{\mathcal{S}_{c}}^{i} \text{log}  (G_c(\Phi(\mathcal{X}_{\mathcal{S}}^{i}))), 
\end{equation}
where $\mathcal{Y}_{\mathcal{S}_{c}}^{i} \in [0, 1]^{C}$ is the binary indicator of each class $c$ in true label for observation $\Phi(\mathcal{X}_{\mathcal{S}}^{i})$, and $G_c(\Phi(\mathcal{X}_{\mathcal{S}}^{i}))$ is the predicted probability of class $c$ (using the softmax function as shown in Fig.~\ref{fig:model}).

\subsection{Kernel for Gaussian distribution}\label{sec:kgd}
We assume that the batch-wise features vector $B_\mathcal{Z}$ (which is one batch data of $G(\Phi(\mathcal{X_Z}))$) follows a Gaussian distribution $\mathcal{N}(\mu_\mathcal{Z}, \Sigma_\mathcal{Z})$, where $\mathcal{Z}$ can be either the source or target domain, and $\mu_\mathcal{Z}$ and $\Sigma_\mathcal{Z}$ are the data mean and covariance respectively:
\begin{equation}
    \mu_\mathcal{Z} = \frac{1}{\mathcal{N}_B} \sum_{z=1}^{\mathcal{N}_B} B_\mathcal{Z}^z
\end{equation}
\begin{equation}
    \Sigma_\mathcal{Z} = \frac{1}{\mathcal{N}_B} \sum_{z=1}^{\mathcal{N}_B} (B_\mathcal{Z}^z - \mu_\mathcal{Z}) (B_\mathcal{Z}^z - \mu_\mathcal{Z})^T
\end{equation}
Therefore, we can calculate the batch-wise covariance matrix of source $\Sigma_\mathcal{S}$ and target  $\Sigma_\mathcal{T}$ domain, respectively. The Gaussian jointly considers the first-order statistic mean and second-order statistic covariance in one single model. Then a form of RBF kernel can be denoted as: 
\begin{equation}\label{eq:kernel}
    \mathcal{K} =  exp (-\kappa ||\Sigma_\mathcal{S} - \Sigma_\mathcal{T} ||_F^2),
\end{equation}
where $||\cdot||_F^2$ is the Frobenius norm. This differs from previous work~\cite{sun2016deep}, which only minimizes the difference between covariance matrices between two domains and reduces Log-Euclidean distance loss. These existing loss functions are largely based on the covariance matrices while ignoring original data. To alleviate this issue, we define the spherical manifold Gaussian kernel geodesic loss to incorporate both covariance matrices and original data.

\subsection{Spherical manifold Gaussian kernel geodesic loss}
By combining the defined Gaussian kernel (Eq.~\ref{eq:kernel}) and batch-wise feature vectors ($B_\mathcal{Z}$), we can measure the intrinsic distance between two domains on one underlying Riemannian spherical manifold.  Therefore, we first project two subspaces into a spherical manifold as follows.
\begin{equation}
\begin{aligned}
     & p = \phi_{Proj.} ( B_\mathcal{S} \times \mathcal{K} ) \\ &
     q = \phi_{Proj.} ( B_\mathcal{T} \times \mathcal{K}  ), 
\end{aligned}
\end{equation}
where $\phi_{Proj.}$ projects data into a $|C|^2$ dimensional spherical manifold and is defined as $\phi_{Proj.} (x) = x.\text{reshape}(-1) / norm(x) $. It first reshapes data into a  $|C|^2$ dimensional space ($|C|$ is the number of categories) and then projects data into a unit  $|C|^2$ dimensional sphere as shown in Fig.~\ref{fig:model} (for better visualization, we only show a 3D sphere). Therefore, we can define the spherical manifold Gaussian kernel geodesic loss as:  
\begin{equation}\label{eq:k}
    \mathcal{L_K} = || \Log_p(q) ||_F^2.
\end{equation}
$\mathcal{L_K}$ can estimate the true geodesic distance between two points on the sphere with the closed-form solution in Eq.~\ref{eq:Log_Sphere}. During the training,  minimizing this loss function directly leads to a small distance between the source and target domains, which is equivalent to minimizing the marginal distribution between two domains. 

\subsection{Conditional distribution alignment}
Since there are no labels in the target domain, we first generate the pseudo labels for the target domain. To mitigate the detrimental effects of bad pseudo-labels, we employ a $T$ times easy-to-hard pseudo-label refinement process to improve the quality of the pseudo-labels in the target domain. Given the trained classifier $G$ in Eq.~\ref{eq:lc}, we can get predicted probability for each target sample as $\text{Softmax} (G(\Phi(\mathcal{X}_{\mathcal{T}}^{j})))$, and the dominate class label is $\mathcal{Y}_\mathcal{PT}^j = \text{max}\ \text{Softmax} (G(\Phi(\mathcal{X}_{\mathcal{T}}^{j})))$ . Easy samples are those whose dominant predicted class probability is bigger than a certain threshold $P_t$. Therefore, one easy pseudo labeled target example can be defined as:
\begin{equation}
    (\Phi(\mathcal{X}_{\mathcal{PT}}^{j}), \mathcal{Y}_\mathcal{PT}^j) \ \text{if} \ \text{max} (\text{Softmax} (G(\Phi(\mathcal{X}_{\mathcal{T}}^{j})))) >  P_t
\end{equation}
During $T$ times easy-to-hard pseudo-label refinement process, for easy examples, $P_t$ has a higher value and for hard examples, $P_t$ has a lower value, hence $P_1 > P_2 > \cdots > P_T$. We define the pseudo labeled target domain as: $\mathcal{D_{PT}} = \{\Phi(\mathcal{X}_\mathcal{PT}^j), \mathcal{Y}^j_\mathcal{PT} \}_{j=1}^\mathcal{N_{PT}}$ with $\mathcal{N_{PT}}$ samples. 

After obtaining the pseudo labeled target domain, we can first optimize the shared classifier $G$ with pseudo labeled target samples in Eq.~\ref{eq:tlc}.
\begin{equation}\label{eq:tlc}
     \mathcal{L_T} = - \frac{1}{\mathcal{N}_\mathcal{PT}}\sum_{j=1}^{\mathcal{N}_\mathcal{PT}} \sum_{c=1}^{C} \mathcal{Y}_{\mathcal{PT}_{c}}^{j} \text{log}  (G_c(\Phi(\mathcal{X}_{\mathcal{PT}}^{j})))   
\end{equation}
For the conditional distribution alignment, differing from Sec.~\ref{sec:kgd}, we use categorical batch-wise feature vectors $\mathcal{C} (B_\mathcal{Z})$ instead of $B_\mathcal{Z}$ since we have labels for both the source and the target domain. $\mathcal{C} (B_\mathcal{Z})$ is the categorical data, which consists of all data in same categories. For example, if the category is 1, then $\mathcal{C} (B_\mathcal{S}) = B_\mathcal{S}[B_\mathcal{{Y_S}}==1]$, and $\mathcal{C} (B_\mathcal{T}) = B_\mathcal{T}[B_\mathcal{{Y_{PT}}}==1]$.  We again assume that $\mathcal{C} (B_\mathcal{Z})$ follows a Gaussian distribution $\mathcal{N}(\mathcal{C}(\mu_\mathcal{Z}), \mathcal{C}(\Sigma_\mathcal{Z}))$, where $\mathcal{Z}$ can be either source or the target domain, and $\mathcal{C}(\mu_\mathcal{Z})$ and $\mathcal{C}(\Sigma_\mathcal{Z})$ are the  mean and covariance of $\mathcal{C} (B_\mathcal{Z})$:
\begin{equation}
    \mathcal{C}(\mu_\mathcal{Z}) = \frac{1}{\mathcal{N}_{\mathcal{C}(B)}} \sum_{z=1}^{\mathcal{N}_{\mathcal{C}(B)}} \mathcal{C}(B_\mathcal{Z}^z)
\end{equation}
\begin{equation}
    \mathcal{C}(\Sigma_\mathcal{Z}) = \frac{1}{\mathcal{N}_{\mathcal{C}(B)}} \sum_{z=1}^{\mathcal{N}_{\mathcal{C}(B)}} ( \mathcal{C}(B_\mathcal{Z}^z) - \mathcal{C}(\mu_\mathcal{Z})) (\mathcal{C}(B_\mathcal{Z}^z) - \mathcal{C}(\mu_\mathcal{Z}))^T
\end{equation}
Therefore, the categorical RBF kernel can be denoted as: 
\begin{equation}\label{eq:ckernel}
    \mathcal{C}(\mathcal{K}) =  exp ( -\kappa ||\mathcal{C}(\Sigma_\mathcal{S}) - \mathcal{C}(\Sigma_\mathcal{T}) ||_F^2),
\end{equation}  

The categorical spherical manifold Gaussian kernel geodesic loss is defined as:
\begin{equation}\label{eq:ck}
    \mathcal{L}_\mathcal{K}^c = || \Log_{\phi_{Proj.} ( \mathcal{C}(B_\mathcal{S}) \times \mathcal{C}(\mathcal{K}) )} ( \phi_{Proj.} ( \mathcal{C}(B_\mathcal{T}) \times \mathcal{C}(\mathcal{K}) ) ) ||_F^2.
\end{equation}
During the training, minimizing $\mathcal{L}_\mathcal{K}^c$ can lead to categorical features between source and target domains to be close to each other. We hence can align the conditional distribution between two domains. 

\subsection{DSGK model}
The framework of our proposed DSGK model is depicted in Fig.~\ref{fig:model}.  Taken altogether, our model minimizes the following objective function:
\begin{equation}\label{eq:loss_all}
  \mathop{\argmin}  \   (\mathcal{L_S}   + \mathcal{L_T} + \alpha \mathcal{L_{K}}  + \beta\frac{1}{C} \sum_{c=1}^{C} \mathcal{L}_\mathcal{K}^c )    
\end{equation}
where $\mathcal{L_S}$ is the source classification loss and  $\mathcal{L_{K}}$ minimizes marginal discrepancy between two domains. $\mathcal{L_T}$ is the pseudo labeled target domain classification loss, and $\mathcal{L}_\mathcal{K}^c$ is the loss function  reducing the conditional categorical discrepancy between two domains. $\mathcal{L_T}$  is equally important as $\mathcal{L_S}$ since we treat pseudo labels as real target labels. The overall training algorithm is shown in Alg.~\ref{alg:DSGK}.

\begin{algorithm}[h]
   \caption{Deep Spherical Manifold Gaussian Kernel Network. $B_\mathcal{S}$ and $B_\mathcal{T}$ denote the mini-batch training sets, $I$ is number of iterations. $T$ is the number  refinement steps.}
   \label{alg:DSGK}
\begin{algorithmic}[1]
   \STATE {\bfseries Input:} labeled source samples  $\mathcal{D_S} = \{\mathcal{X}_\mathcal{S}^i, \mathcal{Y}_\mathcal{S}^i \}_{i=1}^{\mathcal{N}_\mathcal{S}}$ and unlabeled target samples $\mathcal{D_T} = \{\mathcal{X}_\mathcal{T}^j\}_{j=1}^{\mathcal{N}_\mathcal{T}}$ 
   \STATE {\bfseries Output:} predicted target domain labels 
   \REPEAT
   \STATE Derive $B_\mathcal{S}$ and $B_\mathcal{T}$ sampled from $\mathcal{D_S}$ and $\mathcal{D_T}$
   \STATE Initialize $\Phi$ and $G$ using Eqs.~\ref{eq:lc} and~\ref{eq:k}, output: $G$
   \FOR{$t=1$ {\bfseries to} $T$}
   \FOR{$i=1$ {\bfseries to} $I$}
   \STATE Generate pseudo-labels $\mathcal{Y}_\mathcal{T_P}$ using $G$ 
   \STATE Derive $\hat{B_\mathcal{T}}$ sampled from pseudo-labeled $\mathcal{D_{PT}} = \{\Phi(\mathcal{X}_\mathcal{PT}^j), \mathcal{Y}^j_\mathcal{PT} \}_{j=1}^\mathcal{N_{PT}}$ 
   \STATE Perform conditional distribution alignment using Eqs.~\ref{eq:tlc} and~\ref{eq:ck}
   \STATE Refine $G$ using Eq.~\ref{eq:loss_all}
   \ENDFOR
   \ENDFOR
   \UNTIL{converged}
\end{algorithmic}
\end{algorithm}
\vspace{-0.3cm}
\subsection{Theoretical Analysis}\label{sec:the}
In this section, we theoretically show the error bound of the target domain for our proposed DSGK model with the domain adaptation theory~\cite{ben2007analysis} in Theorem 1. 

\textbf{Theorem 1} \textit{Let $\mathcal{H}$ be a hypothesis space.  Given two domains $\mathcal{D_S}$ and $\mathcal{D_T}$,  we have
\begin{equation*}
\begin{aligned}\label{eq:ana}
\forall h \in \mathcal{H},\  R_{\mathcal{T}} (h)  & 
\leq R_{\mathcal{S}} (h) + d_{\mathcal{H}\Delta\mathcal{H}}(\mathcal{D_S}, \mathcal{D_T}) + \beta, 
\vspace{-0.3cm}
\end{aligned}
\end{equation*}
where $R_{\mathcal{S}} (h)$ and $R_{\mathcal{T}} (h)$ represent the source and target domain risk, respectively. $d_{\mathcal{H}\Delta\mathcal{H}}$ is the discrepancy distance between two distributions $\mathcal{D_S}$ and $\mathcal{D_T}$ (including both marginal and conditional distributions). $\beta = \argmin_{h\in\mathcal{H}} R_{\mathcal{S}} (h^{*}, f_\mathcal{S}) + R_{\mathcal{T}} (h^{*}, f_\mathcal{T})$  where $f_\mathcal{S}$ and $f_\mathcal{T}$ are the label functions of two domains, which can be determined by $\mathcal{Y_S}$ and pseudo target domain labels. $h^*$ is the ideal hypothesis and $\beta$ is the shared error and is expected to be negligibly small and can be disregarded.}

In our DSGK model, the first term $R_{\mathcal{S}} (h)$ can be small by training the labeled source domain in Eq.~\ref{eq:lc}. During the training, the domain discrepancy distance $d_{\mathcal{H}\Delta\mathcal{H}}$ can be minimized by reducing the divergence between the marginal and target distributions of latent feature space of two domains. Specifically, $d_{\mathcal{H}\Delta\mathcal{H}} \approx \mathcal{L_K} + \frac{1}{C} \sum_{c=1}^{C} \mathcal{L}_\mathcal{K}^c$.  Ideally, the domain discrepancy distance will be perfectly removed if $\mathcal{L_K} + \frac{1}{C} \sum_{c=1}^{C} \mathcal{L}_\mathcal{K}^c$ is close to 0. However, it can be achieved if and only if  $\mathcal{D_S} = \mathcal{D_T}$. Therefore, minimizing $d_{\mathcal{H}\Delta\mathcal{H}}$ is  equivalent to minimizing $\mathcal{L_K} + \frac{1}{C} \sum_{c=1}^{C} \mathcal{L}_\mathcal{K}^c$. 

\section{Experiments}
\subsection{Setup}
\begin{figure}[h]
\centering
\includegraphics[width=1.0\columnwidth]{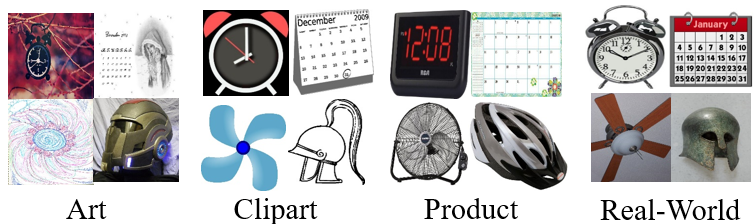}
\caption{Sample images from four domains of the Office-Home dataset. We only show images from four categories.}
\vspace{-0.6cm}
\label{fig:oh}
\end{figure}

 \paragraph{Datasets.}We test our model using three image datasets:  Office-31, Office-Home, and VisDA-2017. \textbf{Office-31} \cite{saenko2010adapting} has 4,110 images from three  domains: Amazon (A), Webcam (W), and DSLR (D) in 31 classes. In experiments, A$\shortrightarrow$W represents transferring knowledge from domain A to domain W.  \textbf{Office-Home}~\cite{venkateswara2017deep} dataset contains 15,588 images from four domains: Art (Ar), Clipart (Cl), Product (Pr), and Real-World (Rw) in 65 classes. Fig.~\ref{fig:oh} shows some example images of four domains.  \textbf{VisDA-2017}~\cite{peng2017visda} is a challenging dataset due to the big domain-shift between the synthetic images (152,397 images from VisDA) and the real images (55,388 images from COCO) in 12 classes. We test our model on the setting of synthetic-to-real as the source-to-target domain and report the accuracy of each category. 
 
\begin{table*}[h!]
\begin{center}
\small
\caption{Accuracy (\%) on Office-Home dataset (based on ResNet50)}
\vspace{-0.3cm}
\setlength{\tabcolsep}{+1.1mm}{
\begin{tabular}{rccccccccccccc}
\hline \label{tab:OH}
Task & Ar$\shortrightarrow$Cl &  Ar$\shortrightarrow$Pr & Ar$\shortrightarrow$Rw & Cl$\shortrightarrow$Ar & Cl$\shortrightarrow$Pr & Cl$\shortrightarrow$Rw & Pr$\shortrightarrow$Ar & Pr$\shortrightarrow$Cl & Pr$\shortrightarrow$Rw & Rw$\shortrightarrow$Ar & Rw$\shortrightarrow$Cl & Rw$\shortrightarrow$Pr & \textbf{Ave.}\\
\hline
ResNet-50~\cite{he2016deep}&	34.9&	50.0&	58.0&	37.4&	41.9&	46.2&	38.5&	31.2&	60.4&	53.9&	41.2&	59.9&	46.1\\
DAN~\cite{long2015learning}	& 43.6	& 57.0& 	67.9& 	45.8& 	56.5& 	60.4& 	44.0& 	43.6& 	67.7& 	63.1& 	51.5& 	74.3& 	56.3\\
DANN~\cite{ghifary2014domain} 	& 45.6	& 59.3& 	70.1& 	47.0& 	58.5& 	60.9& 	46.1& 	43.7& 	68.5& 	63.2& 	51.8& 	76.8& 	57.6\\
JAN~\cite{long2017deep}		& 45.9& 	61.2& 	68.9& 	50.4& 	59.7& 	61.0& 	45.8& 	43.4& 	70.3& 	63.9& 	52.4& 	76.8& 	58.3\\
CDAN-M~\cite{long2018conditional}	& 50.6& 	65.9& 	73.4& 	55.7& 	62.7& 	64.2& 	51.8& 	49.1& 	74.5& 	68.2& 	56.9& 	80.7& 	62.8\\
TAT~\cite{liu2019transferable} &51.6  &69.5 & 75.4 & 59.4 & 69.5 & 68.6 & 59.5 &50.5 &76.8 &70.9 &56.6 &81.6 &65.8 \\
ETD~\cite{li2020enhanced} &51.3 & 71.9& 85.7& 57.6 &69.2 &73.7 &57.8 &51.2 &79.3 &70.2 &57.5 &82.1 &67.3 \\
TADA~\cite{wang2019transferable} & 53.1 &72.3& 77.2& 59.1 &71.2& 72.1& 59.7& 53.1& 78.4 &72.4& \textbf{60.0} &82.9& 67.6 \\
SymNets~\cite{zhang2019domain} & 47.7 & 72.9 & 78.5 & 64.2  & 71.3  &74.2  & 64.2  & 48.8 &  79.5&  74.5 &52.6 & 82.7& 67.6 \\
DMP~\cite{luo2020unsupervised} & 52.3 &73.0 &77.3 &64.3 &72.0 &71.8 &63.6 &52.7 &78.5 &72.0 &57.7 &81.6 & 68.1 \\
DCAN~\cite{li2020domain} & 54.5 &75.7 &81.2 &67.4 &74.0 &76.3 &67.4 &52.7 &80.6 &74.1 &59.1 &83.5 &70.5\\ 
\hline
\hline
\textbf{DSGK}& \textbf{55.9 } &	\textbf{78.4}  &	\textbf{81.3}  &	\textbf{69.1} &	\textbf{81.9}  & \textbf{80.2} & \textbf{70.1} &	\textbf{55.7} &	\textbf{82.1} &	\textbf{75.1} &	58.4  &	\textbf{84.9} & 	\textbf{72.8} \\
\hline
\end{tabular}}
\vspace{-0.3cm}
\end{center}
\end{table*}

\begin{table*}[h!]
\begin{center}
\small
\caption{Accuracy (\%) on VisDA-2017 dataset (based on ResNet101)}
\vspace{-0.3cm}
\setlength{\tabcolsep}{+2.1mm}{
\begin{tabular}{rccccccccccccc}
\hline \label{tab:VisDA}
Task & plane& bcycl& bus& car& horse& knife& mcycl& person& plant& sktbrd& train& truck & \textbf{Ave.}\\
\hline
Source-only~\cite{he2016deep} &  55.1 &53.3 &61.9& 59.1& 80.6& 17.9& 79.7& 31.2& 81.0& 26.5& 73.5& 8.5 & 52.4 \\
DANN~\cite{ghifary2014domain} 	&81.9 &77.7& 82.8 &44.3& 81.2& 29.5 &65.1 &28.6 & 51.9 & 54.6 & 82.8 & 7.8 & 57.4\\
DAN~\cite{long2015learning}	& 87.1& 63.0& 76.5& 42.0 &90.3& 42.9 &85.9 &53.1& 49.7 &36.3& 85.8 &20.7 &61.1\\
JAN~\cite{long2017deep}	& 75.7& 18.7& 82.3 &86.3& 70.2 &56.9& 80.5& 53.8 &92.5 &32.2& 84.5& 54.5 &65.7 \\
MCD~\cite{saito2018maximum}	& 87.0 &60.9& 83.7& 64.0& 88.9& 79.6& 84.7& 76.9& 88.6& 40.3& 83.0& 25.8& 71.9\\
DMP~\cite{luo2020unsupervised} &92.1 &75.0 &78.9 &75.5 &91.2 &81.9 &89.0 &77.2 &93.3 &77.4 &84.8 &35.1 &79.3 \\
DADA~\cite{tang2020discriminative} & 92.9 &74.2& 82.5& 65.0& 90.9& 93.8& 87.2& 74.2& 89.9& 71.5& 86.5 &48.7&  79.8 \\
STAR~\cite{lu2020stochastic} & 95.0& 84.0& 84.6& 73.0& 91.6 &91.8& 85.9 &78.4& 94.4& 84.7 &87.0 &42.2& 82.7 \\
\hline
\hline
\textbf{DSGK}&   \textbf{95.7} & \textbf{86.3} & \textbf{85.8} & \textbf{77.3} & \textbf{92.3} & \textbf{94.9} & \textbf{88.5} & \textbf{82.9} & \textbf{94.9} & \textbf{86.5} & \textbf{88.1} & \textbf{46.8} & \textbf{85.0} \\
\hline
\end{tabular}}
\vspace{-0.3cm}
\end{center}
\end{table*}


\begin{table}[!htb]
\small
      \caption{Accuracy (\%) on Office-31 dataset (based on ResNet50)}
      \vspace{-0.3cm}
      \centering
\setlength{\tabcolsep}{+0.3mm}{
\begin{tabular}{rcccccccc|c|c|c|c|c|c|c|c|}
\hline \label{tab:O31}
Task & A$\shortrightarrow$W &  A$\shortrightarrow$D & W$\shortrightarrow$A & W$\shortrightarrow$D & D$\shortrightarrow$A & D$\shortrightarrow$W  & \textbf{Ave.}\\
\hline
ResNet50~\cite{he2016deep} &  68.4 & 68.9 & 60.7 & 99.3 & 62.5 & 96.7 & 76.1 \\
RTN~\cite{long2016unsupervised} &	84.5 &	77.5 &	64.8 &	99.4 &	66.2 &	96.8 &	81.6 \\
ADDA~\cite{tzeng2017adversarial}	&86.2	&77.8  &68.9	&98.4 &	69.5 &	96.2	& 82.9\\
GSM~\cite{zhang2019transductive}	& 84.8  & 82.7  & 73.5   &96.6  & 70.9  & 95.0  & 83.9\\
JAN~\cite{long2017deep}&	85.4 &	84.7	&70.0 &	99.8	&68.6 &	97.4 &	84.3\\
ETD~\cite{li2020enhanced} &92.1 & 88.0 & 67.8 & \textbf{100} & 71.0 & \textbf{100} &  86.2 \\
DMP~\cite{luo2020unsupervised} & 93.0 & 91.0 & 70.2 & \textbf{100} & 71.4 & 99.0 & 87.4\\
TADA~\cite{wang2019transferable} &94.3 & 91.6  & 73.0  &99.8  & 72.9 & 98.7 & 88.4 \\
SymNets~\cite{zhang2019domain}& 90.8& 93.9  &72.5  & \textbf{100} & 74.6 & 98.8& 88.4 \\
TAT~\cite{liu2019transferable} & 92.5 & 93.2 & 73.1 & \textbf{100}& 73.1 & 99.3 & 88.5 \\
MDA~\cite{zhang2019modified}& 94.0  & 92.6  & 77.6 &  99.2  & 78.7  & 96.9  & 89.8\\
CAN~\cite{kang2019contrastive} & 94.5 & 95.0  &77.0   &99.8  & 78.0 & 99.1 &  90.6 \\
\hline
\hline
\textbf{DSGK} & \textbf{95.7} &	\textbf{96.9 } &	\textbf{80.3} &	99.6 &	\textbf{80.3} &	99.2 &	\textbf{92.0}	\\
\hline
\end{tabular}}
\vspace{-0.3cm}
\end{table}

\begin{table}[b]
\vspace{-0.4cm}
\small
\begin{center}
\caption{Ablation experiments on Office-31 dataset
\label{tab:ablation}}
\vspace{-0.3cm}
\setlength{\tabcolsep}{+0.6mm}{
\begin{tabular}{lcccccccc}
\hline \label{tab:ab}
Task & A$\shortrightarrow$W &  A$\shortrightarrow$D & W$\shortrightarrow$A & W$\shortrightarrow$D & D$\shortrightarrow$A & D$\shortrightarrow$W  & \textbf{Ave.}\\
\hline
DSGK$-$K/T/C	& 85.2 &	89.2 &	75.0 &	97.6 &	75.4 &	94.7 &	86.2\\
DSGK$-$C/T  & 85.9 &	90.1 &	75.2&	98.0 &	76.0 &	96.1&	86.9\\
DSGK$-$K/T &91.6 &	90.7 &	78.0  &	98.1 &	77.8 &	96.3 &	88.8\\	
DSGK$-$K/T 	&93.4 &	91.6&	78.0&	98.3&	78.3&	96.9&	89.4\\
DSGK$-$C	& 94.5 &94.3&	78.7&	99.0 &	78.6 &	97.0&	90.4\\
DSGK$-$T & 94.9 &	95.3 &	79.0 &	99.0  &	79.0 &	97.4&	90.8\\
DSGK$-$K & 95.3 &	96.4 &	79.5 &	99.0 &	79.2 &	98.0 &	91.2	\\
\hline
\hline
\textbf{DSGK} & \textbf{95.7} &	\textbf{96.9 } &	\textbf{80.3} &	\textbf{99.6} &	\textbf{80.3} &	\textbf{99.2 }&	\textbf{92.0} \\  
\hline
\end{tabular}}
\end{center}
\end{table}

\paragraph{Implementation details.}
We implement our approach using PyTorch and extract features for the three datasets from  finely tuned ResNet50 (Office-31, Office-Home) and ResNet101 (VisDA-2017) networks~\cite{he2016deep}. The 1,000 features are then extracted from the last fully connected layer for the source and target features. The output of three Linear layers are 512, 256 and $|C|$, respectively. Parameters in recurrent pseudo labeling are  $T=5$ and $P_T=[0.9, 0.8, 0.7, 0.6, 0.5]$. Learning rate ($\epsilon = 0.001$), batch size (64), $\kappa = 0.1$, $\alpha =0.1$, $\beta =0.01$ and number of epochs (9) are determined by performance on the source domain.  We compare our results with 20 state-of-the-art methods\footnote{Source code is available at: \url{https://github.com/YoushanZhang/Transfer-Learning/tree/main/Code/Deep/DSGK}.}. Experiments are performed with a GeForce 1080 Ti.

\subsection{Results}

\begin{figure*}[t]
\centering
\begin{subfigure}{0.42\textwidth}
\includegraphics[width=\linewidth]{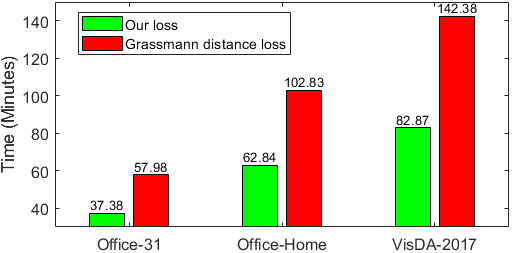}
\caption{Time of three datasets  } \label{fig:time_all}
\end{subfigure}
\begin{subfigure}{0.42\textwidth}
\includegraphics[width=\linewidth]{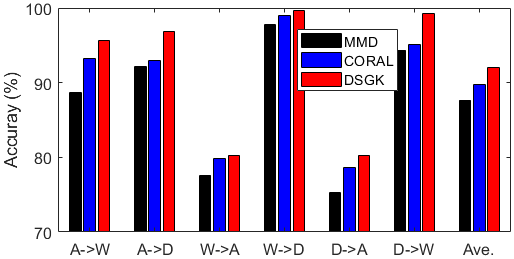}
\caption{Accuracy comparison of different loss functions} \label{fig:time_o31}
\end{subfigure} 
\vspace{-0.2cm}
\caption{Computation time and different loss functions comparison. (a) is the total computation time of all transfer tasks in three datasets (six for Office-31, twelve for Office-Home and one for VisDA-2017 dataset ). On average, our DSGK model is approximately 1.6 times faster than Grassmann distance loss. (b) compares the DSGK model with the other two loss functions in each task of Office-31. Our loss function achieves a higher accuracy than the other two.   } \label{fig:o31}
\vspace{-0.4cm} 
\end{figure*}

The performance on Office-Home,  VisDA-2017, and Office-31 are shown in Tables~\ref{tab:OH}-\ref{tab:O31}.  Our DSGK model outperforms all state-of-the-art methods in terms of average accuracy (especially in the VisDA-2017 and Office-Home datasets). The DSGK model substantially improves classification accuracy on difficult adaptation tasks (e.g., W$\shortrightarrow$A task in the Office-31 dataset and the challenging VisDA-2017 and Office-Home datasets, which have a larger number of categories and different domains are visually dissimilar).
   
In the Office-31 dataset, the mean accuracy is 92.0\%, compared with the best baseline (CAN), our DSGK model provides a 1.4\%  improvement. Although the improvement is not significant, we have an obvious improvement in some difficult tasks (W$\shortrightarrow$A and D$\shortrightarrow$A). The mean accuracy on the Office-Home dataset is increased from 70.5\% (DCAN) to 72.8\%. We also notice that accuracy is obviously improved across all tasks except Pr$\shortrightarrow$Cl.  In the VisDA-2017 dataset, the DSGK model has a 2.3\% improvement over the best baseline (STAR), and it achieves the highest performance in all tasks. Therefore, our proposed spherical manifold Gaussian kernel loss is useful, and the easy-to-hard refinement process is effective in improving the classification accuracy.

In addition, we compare the computation time of our proposed DSGK model with Grassmann distance in DMP, which relies on the SVD of the covariance matrices in Fig.~\ref{fig:time_all}. DSGK model requires relatively less computation time for all three datasets. Our loss functions are (1.5, 1.6, and 1.7 times) faster than Grassmann distance loss in the three datasets since Grassmann distance loss requires the calculation-intensive SVD. Therefore, our proposed loss function is much faster than Grassmann distance loss. To show the effectiveness of the proposed spherical manifold Gaussian kernel geodesic loss,  we also compare the results of well-used loss functions: CORAL loss~\cite{sun2016deep} and MMD~\cite{long2015learning} loss; that is, we replace Eq.~\ref{eq:k} and Eq.~\ref{eq:ck} with these two loss functions.  As shown in Fig.~\ref{fig:o31}, our proposed loss achieves higher accuracy than CORAL and MMD loss functions.  Therefore, our DSGK model is effective and accurate in UDA tasks. 

\subsection{Ablation study}
To demonstrate the effects of different loss functions on final classification accuracy, we present an ablation study in Tab.~\ref{tab:ablation}, in which K represents  spherical manifold  Gaussian kernel geodesic loss $\mathcal{L_K}$, T is the $\mathcal{L_T}$ and C denotes the $\mathcal{L}_\mathcal{K}^c$. ``DSGK$-$K/T/C” is implemented without $\mathcal{L_K}$, $\mathcal{L_T}$, and $\mathcal{L}_\mathcal{K}^c$. It is a simple model, which only reduces the source risk without minimizing the domain discrepancy using $\mathcal{L_S}$. ``DSGK$-$C/T” only aligns the marginal distribution between two domains. ``DSGK$-$C” reports results without performing the additional categorical conditional distribution alignment. We observe that with the increasing number of loss functions, the robustness of our model keeps improving. The usefulness of loss functions is ordered as $\mathcal{L_K}<\mathcal{L_{T}}<\mathcal{L}_\mathcal{K}^c$. Therefore, the proposed spherical manifold  Gaussian kernel geodesic loss and easy-to-hard learning approach are effective in improving performance, and different loss functions are helpful and important in minimizing target domain risk.

\subsection{Parameter Analysis}
\begin{figure*}[t]
\centering
\begin{subfigure}{0.48\textwidth}
\includegraphics[width=\linewidth]{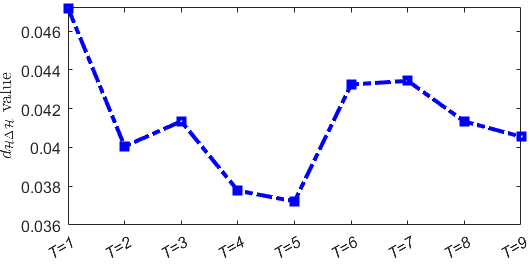}
\caption{Effect of different $T$ on $d_{\mathcal{H}\Delta\mathcal{H}}$} \label{fig:imb}
\end{subfigure} 
\begin{subfigure}{0.51\textwidth}
\includegraphics[width=\linewidth]{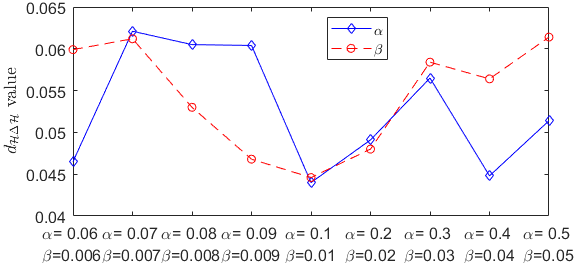}
\caption{Effect of different $\alpha$ and $\beta$ on $d_{\mathcal{H}\Delta\mathcal{H}}$ } \label{fig:ima}
\end{subfigure}
\vspace{-0.1cm}
\caption{Parameter analysis. In (a), $d_{\mathcal{H}\Delta\mathcal{H}}$ is minimum when $T=5$. In (b), the x-axis denotes different $\alpha$ and $\beta$, $d_{\mathcal{H}\Delta\mathcal{H}}$ is minimum when $\alpha = 0.1$ and $\beta = 0.01$.  } \label{fig:rela}
\vspace{-0.3cm}
\end{figure*}
\begin{figure*}[h!]
\centering
\includegraphics[width=2\columnwidth]{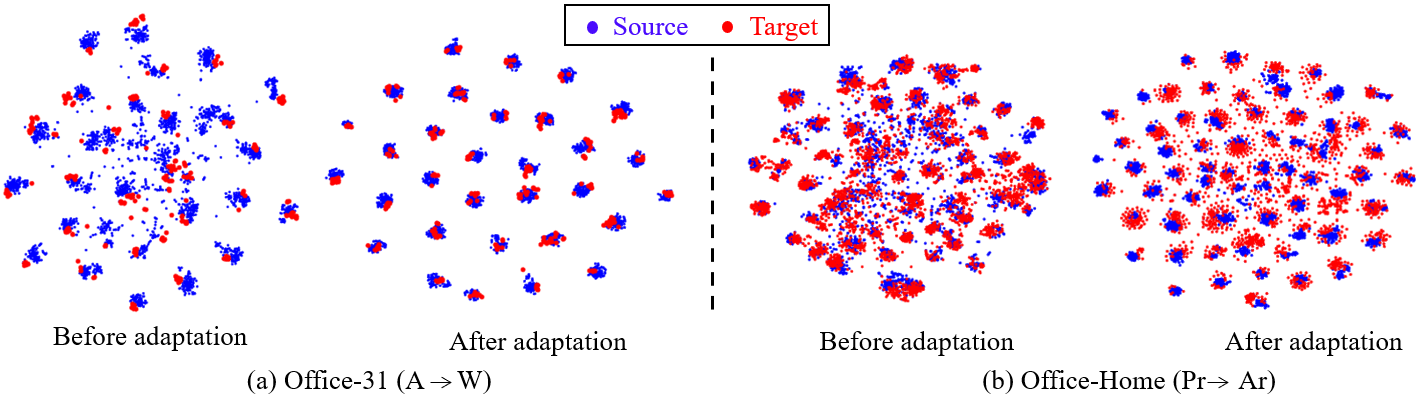}
\caption{Feature visualization using a 2D t-SNE view of task A$\shortrightarrow$W in Office-31 dataset and Pr$\shortrightarrow$Ar in Office-Home dataset.  Our DSGK model improves the discriminative representations across domains. (blue color: source domain, red color: target domain). Best viewed in color.}
\vspace{-0.3cm}
\label{fig:tsne}
\end{figure*}
There are four hyperparameters $T$, $P_t$, $\alpha$ and $\beta$ in DSGK that can affect the final accuracy. To get the optimal parameters, we randomly select the task W$\shortrightarrow$A in Office-31 dataset and run a set of experiments regarding different values of each parameter. Notice that it is inappropriate to tune parameters using the target domain accuracy since we do not have any labels in the target domain. Therefore, we report the $\mathcal{H}$ divergence between two domains, as stated in Sec.~\ref{sec:the}.  Since $\mathcal{H}$ divergence can be assessed by $d_{\mathcal{H}\Delta\mathcal{H}} \approx \mathcal{L_K} + \frac{1}{C} \sum_{c=1}^{C} \mathcal{L}_\mathcal{K}^c$, we can calculate $\mathcal{L_K} + \frac{1}{C} \sum_{c=1}^{C} \mathcal{L}_\mathcal{K}^c$ and select the parameters if they achieve the minimal value. $\alpha$ is selected from $\{0.06, 0.07, 0.08, 0.09, 0.1, 0.2, 0.3, 0.4, 0.5\}$, $\beta$ is selected from $\{0.006, 0.007, 0.008, 0.009, 0.01, 0.02, 0.03, 0.04, 0.05\}$, $T$ is selected from $\{1,2,3,4,5,6,7,8,9\}$, and $P_t$ is selected from $\{0.9,0.8,0.7,0.6,0.5,0.5,0.3,0.2,0.1\}$, we vary one parameter and fix the others at a time. For $T$ and $P_t$, if $T = 9$, then  $P_t = [ 0.9,0.8,0.7,0.6,0.5,0.5,0.3,0.2,0.1]$ and  if $T = 1$, then  $P_t = [0.1]$. 

Fig.~\ref{fig:rela} demonstrates that our DSGK model is not very sensitive to a wide range of parameter values since the $\mathcal{H}$ divergence ($d_{\mathcal{H}\Delta\mathcal{H}}$) is not significantly changed. We first determine the $T$ and $P_t$ as shown in Fig.~\ref{fig:imb}. We can find that when $T =5$, that is $P_t = [0.9, 0.8, 0.7, 0.6, 0.5]$, $d_{\mathcal{H}\Delta\mathcal{H}}$ achieves the minimum value. Hence, $P_t = [0.9, 0.8, 0.7, 0.6, 0.5]$ is the best parameter for our DSGK model. In Fig.~\ref{fig:rela}, a large $T$ can have a larger $d_{\mathcal{H}\Delta\mathcal{H}}$ (e.g., $T = 6$ in Fig.~\ref{fig:imb} and $T = 7$ in Fig.~\ref{fig:imb}) since a larger $T$ brings hard examples into the shared classifier $G$, which leads to lower perforamce in the target domain. Therefore, the parameter analysis is useful in finding the best hyperparameters for our DSGK model.  After fixing $T$ and $P_t$, in Fig.~\ref{fig:ima}, we combine the parameter tuning results for $\alpha$ and $\beta$ together. If it is $\alpha$, then the x-axis is from 0.1 to 0.9, and if it is $\beta$, then the x-axis is from 0.01 to 0.09. It shows that when $\alpha = 0.1$ and $\beta =0.01$ achieves the minimum number. Therefore, the hyperparameter $\alpha = 0.1$ and $\beta =0.01$ is the best since the discrepancy between two domains is minimized.

\subsection{Feature Visualization}
To intuitively present adaptation performance during the transition from the source domain to the target domain, we utilize-SNE to visualize the deep features of network activations in 2D space before and after distribution adaptation. Fig.~\ref{fig:tsne} visualizes embeddings of the task A$\shortrightarrow$W in the Office-31 dataset and Pr$\shortrightarrow$Ar in the Office-Home dataset. We can observe that the representation becomes more discriminative after adaptation, while many categories are mixed in the feature space before adaptation. Therefore, DSGK can learn more discriminative representations, which can significantly increase inter-class dispersion and intra-class compactness.

\section{Discussion}
In these experiments, DSGK always achieves the highest average accuracy. Therefore, the quality of our model exceeds that of SOTA methods and is better than existing loss functions. There are two compelling reasons. First, the proposed spherical manifold  Gaussian kernel geodesic loss can project data into a spherical manifold and calculate the intrinsic distance between two domains. It not only avoids the complex SVD calculation to reduce computation time as in the Grassmannian manifold but also considers both the batch-wise features $B_\mathcal{Z}$ and the Gaussian kernel $\mathcal{K}$ between two domains.  Secondly, to minimize the conditional distribution discrepancy, we develop an easy-to-hard refinement process by keeping reducing the predicted probability of the target domain. This strategy can push the shared classifier $G$ towards the target domain. Hence, the easy-to-hard refinement process is useful in updating the network parameters, which further reduces the domain discrepancy. 

One limitation of DSGK is the 
loss $\mathcal{L}_\mathcal{K}^c$ needs to calculate the difference between two domains of each category. If the number of categories ($|C|$) is substantially larger, it may need more computation time.  However, it will still be faster than existing Grassmann distance loss function, as shown in Fig.~\ref{fig:time_all}.

\section{Conclusion}
In this paper, we propose a novel deep spherical manifold Gaussian kernel (DSGK) model for unsupervised domain adaptation. To align the marginal distribution between the source and the target domain, we develop a spherical manifold Gaussian kernel geodesic loss to minimize the intrinsic distance between two domains. We also employ an easy-to-hard refinement process to remove the noisy pseudo labels and reduce the categorical  spherical manifold  Gaussian kernel geodesic loss to align the conditional distribution of two domains. Extensive experiments demonstrate that the proposed DSGK model achieves higher accuracy than state-of-the-art domain adaptation methods.

{\small
\bibliographystyle{ieee_fullname}
\bibliography{egbib}
}

\end{document}